\documentclass[10pt]{article}

\usepackage[a4paper, total={6.5in, 9in}]{geometry}
\AtBeginDocument{%
  }

\usepackage{url}

\title{A Community-driven vision for a new Knowledge Resource for AI}

\author{Vinay K Chaudhri,
Chaitan Baru,
Brandon Bennett,
Mehul Bhatt, 
Darion Cassel, \\
Anthony G Cohn,
Rina Dechter,
Esra Erdem,
Dave Ferrucci,\\
Ken Forbus, 
Gregory Gelfond,
Michael Genesereth,
Andrew S. Gordon, \\
Benjamin Grosof,
Gopal Gupta,
Jim Hendler,
Sharat  Israni,\\
Tyler R. Josephson,
Patrick Kyllonen,
Yuliya Lierler,
Vladimir Lifschitz,\\
Clifton McFate,
Hande Küçük McGinty,
Leora Morgenstern,\\
Alessandro Oltramari,
Praveen Paritosh,
Dan Roth,
Blake Shepard, \\
Cogan Shimizu,
Denny Vrandečić,
Mark Whiting,
Michael Witbrock }

\begin{document}
\maketitle
\begin{abstract}
The long-standing goal of creating a comprehensive, multi-purpose knowledge resource, reminiscent of the 1984 Cyc project, still persists in AI. Despite the success of knowledge resources like WordNet, ConceptNet, Wolfram\textbar{}Alpha and other commercial knowledge graphs, verifiable, general-purpose widely available sources of knowledge remain a critical deficiency in AI infrastructure. Large language models struggle due to knowledge gaps; robotic planning lacks necessary world knowledge; and the detection of factually false information relies heavily on human expertise. What kind of knowledge resource is most needed in AI today? How can modern technology shape its development and evaluation? A recent AAAI workshop gathered over 50 researchers to explore these questions. This paper synthesizes our findings and outlines a community-driven vision for a new knowledge infrastructure. In addition to leveraging contemporary advances in knowledge representation and reasoning, one promising idea is to build an open engineering framework to exploit knowledge modules effectively within the context of practical applications. Such a framework should include sets of conventions and social structures that are adopted by contributors.
\end{abstract}

\section{Introduction}
The Cyc project, started in 1984, created the first large-scale database of commonsense knowledge. The initiative continues to this day with its aim to provide a comprehensive ontology and knowledge base of commonsense knowledge to enable human-like reasoning for AI systems. 
In the concluding paragraph of his Communications of the Association of Computing Machinery (CACM) 1995 article {\it A Large-Scale Investment in Knowledge Infrastructure}~\cite{lenat1995cyc}, Cyc's founder Douglas B. Lenat wrote:

\begin{quote}
    Is Cyc necessary? How far would a user get with something simpler than Cyc but that lacks everyday commonsense knowledge? Nobody knows; the question will be settled empirically. Our guess is most of these applications will eventually tap the synergy in a suite of sources (including neural nets and decision theory), one of which will be Cyc.    
\end{quote}

Although 30 years have passed since the above article was written, AI research community has not conclusively settled \cite{brachman2022toward} the question “How far would a user get with something simpler than Cyc but that lacks everyday commonsense knowledge?” However, it is clear that significant strides have been made in addressing many of the tasks that were original Cyc use cases, including information retrieval, semi-automatically linking multiple heterogeneous external information sources, spelling and grammar correction, machine translation, natural language understanding and speech understanding. Much of this progress has been facilitated by rapid, profound, and synergistic advancements in neural networks,  
%\cite{aggarwaletal2018},  NOt sure if we need another ciation to add to our already long list.
automated agents, and knowledge graphs \cite{chaudhri2022knowledge}.
%\cite{Ehrlingeretal2016a}.
The increasing importance of knowledge graphs is in line with Lenat's early vision for Cyc, as he presented in 1985 \cite{lenat1985cyc}.
%\cite{lenatandGuha1990}\cite{lenatetal1990}. I went with the earliest paper on Cyc that I could find.

%However, encoding commonsense knowledge does not appear to be a major concern for researchers in these crucial fields. 
% Above statement is not correct because many LLM researchers focus on commonsense - of course not the same way as Cyc does.
Knowledge graphs that have had commercial impact, for example, Google's knowledge graph, have primarily focused on capturing relationships among real-world entities and have sidestepped the challenge of capturing everyday commonsense knowledge. Likewise, even though large language models (LLMs) perform very well on tasks that require knowledge about the world, that knowledge is most often implicit. Furthermore, it has been difficult to make formal guarantees about how and when LLMs leverage this implicit knowledge \cite{aaai2025panel}.

The lack of large-scale commonsense knowledge bases hinders practical AI applications. For example, robotic task and motion planning would significantly benefit from world knowledge, and commonsense task descriptions to be performed by a robot \cite{IROS19-jiang}. At present, such knowledge is custom built for each project. Pressing problems such as identifying factually false information~\cite{nasem2023nobel} require trusted knowledge; without it, systems must resort to guess work, and are forced into over-reliance on human expertise to guide them. To ground the narrative in this paper, the examples of “robotics” and “identifying factually false information” are used to illustrate how many applications suffer from a lack of knowledge.

This raises important questions about AI infrastructure. What kind of knowledge resource is most needed in the modern context? What categories of knowledge should be included in such a resource? How should they be represented? How should they be made accessible to external users and applications? How can recent technological advances be harnessed to create such a knowledge resource? How should such a resource be evaluated?

A workshop at the 2025 Conference for the Advancement of Artificial Intelligence (AAAI) gathered over 50 researchers, including the authors of this paper, to explore these questions \cite{TIKA2025}. This paper is an attempt to synthesize our discussions, and present a community view on the requirements and approach for creating a new knowledge infrastructure. We concluded that building an open engineering infrastructure for disseminating and using knowledge modules in practical applications, leveraging steady advances in knowledge representation and reasoning over the last 40 years, is a promising path forward. This includes establishing sets of conventions and social structures that should be adopted by contributors.

We begin by envisioning the type of knowledge resource required in the current context of artificial intelligence (AI), and then exploring aspects of creating it, including foundational knowledge, domain-specific knowledge, automated reasoning, and evaluation. We also review the current state of education in knowledge representation and reasoning. We conclude by presenting a community view on productive steps towards the creation of this much needed knowledge resource.

\section{Envisioning a Knowledge Resource}

We use the term \textit{knowledge resource} to refer to a body of curated knowledge that can be examined and verified by humans. This nowledge could be formalized in any computational framework including ontologies~\cite{noy2001ontology}, rules~\cite{genesereth2022introduction}, a constraint network~\cite{dechter2003constraint}, a probabilistic causal model~\cite{Pearl-book1}, and even in unambiguous natural language~\cite{logical-english}. We envision the knowledge resource needed for AI by exploring four points: the type of problems that could best be solved by such a knowledge resource, the limits of current practice, what should be done differently, and the likely beneficiaries of such a resource. 

\subsection{Problems solvable by a Knowledge Resource}

Data-driven learning is central to modern AI. But in some cases, curated knowledge can be better. For example, in basic arithmetic, people easily outcompete chatbots such as ChatGPT because they learn how to add and multiply numbers using rules rather than by looking at a large number of examples~\cite{cheng-yu-2023-analyzing}.

Curated knowledge works better than data-driven learning in these scenarios: rules needed to solve the problem are readily available, such as in systems of axioms for qualitative spatial reasoning~\cite{cohn2008qualitative,forbus2019qualitative}; in applications that require high levels of accuracy, transparency, and reliability such as income tax calculations; and in contexts in which clarity and precision are crucial, such as in education.

It has been shown through the qualitative spatial reasoning benchmark, Room Space 100,  that When the number of objects in a scene (n) increases from 3 to 6, GPT-4's accuracy generally declines, for example, for n=3, GPT-4 has an accuracy of 0.55, and for n=6, an accuracy of 0.15 \cite{li2024reframing}. It has also been found that GPT-4 could solve some action-based complex puzzles -- performing better when generating Python code -- but an axiomatic reasoner using a proper action description language solved all puzzles correctly \cite{ishay2025llm+}.  A set of concrete examples where a knowledge resource can supplement the knowledge in large language models has been put out by  Wolfram\textbar{}Alpha \cite{wolfram2023wolframalpha}.  These include ``distance between Chicago and Tokyo'', ``3 to the power 73'', ``circumference of an ellipse with axes 4 and 12'', etc.

% This sentence became problematic as we have not defined reasoning. 
%While some AI systems are capable of superhuman reasoning (e.g. SAT solvers, model checkers), they do so over carefully formulated problems in narrow domains.
A knowledge resource is necessary for us to develop automated reasoning and intelligence that is more human-like, capable of participating in formulation of formal models from everyday inputs and interpreting their results in real-world terms (\textit{model formulation} and \textit{model interpretation}\cite{forbus2019qualitative}). This requirement is supported by a recent survey of AI community in which 
61.8\% of survey participants estimated the
minimal percentage of symbolic AI techniques
required for reaching human-level reasoning to
be at least 50\%  \cite{aaai2025panel}. 

\subsection{Limits of the current practice}

Current knowledge resources such as Wikidata~\cite{vrandevcic2014wikidata} and Google’s knowledge graph~\cite{singhal_2012}  capture relationships between real world entities such as people, places, organization, etc. ConceptNet captures relationships between concepts (for example, a knife is used for cutting)~\cite{speer2017conceptnet}, but the relationships are limited to triples that are awkward, or insufficiently expressive, for capturing general-purpose knowledge. Large Language Models (LLMs) encompass enormous amounts of knowledge implicitly.  This is effective for many applications but their output lacks formal guarantees~\cite{aaai2025panel}. 

Applications such as biomedicine require a much higher level of expressivity. For example, a drug may up-regulate or down-regulate a gene depending on the context~\cite{Unni2022BiolinkModel}.  The knowledge in biomedicine is often below the level of certainty of “established facts.”  In clinical practice, a great deal of reasoning has to rely on assertions, statistical associations and observations.  Existing reasoning engines are unable to accommodate varying levels of certainty at the discretion of the practitioner.

Knowledge resources such as Cyc~\cite{lenat1995cyc}, Wolfram\textbar{}Alpha \cite{Wolfram|One}, Component Library~\cite{barker01} or its variant CoreALMLib~\cite{inclezan16}
attempt to bridge this gap. Cyc, for example, targets formalization of complex knowledge patterns across the full spectrum of the human world. Because of the restricted intellectual property of Cyc and Wolfram\textbar{}Alpha, limited access to them is not conducive to fostering an open-source community.  OpenCyc, which is the open-source version of Cyc, is useful for limited purposes as it provides access to only a small fraction of the full knowledge base \cite{OpenCyc-4.0}.

Demonstration of inference gaps in LLMs for spatial reasoning and action reasoning cited in the previous section have been limited to academic settings.  Establishing that such inference gaps are critical for real-world applications  is an open research problem.

\subsection{What should be different about a modern knowledge resource?}

A modern knowledge resource should provide a public-utility-like infrastructure becoming a go to place for trusted and verified knowledge and reasoning methods across a variety of domains.  Towards that end, it should provide knowledge in such a way that its formal representation is paired with its provenance in multiple modalities including text, images, video, or other formats such as graphs. As most pieces of knowledge are not universally true, emphasis must be placed on the  applicable context of that knowledge. 

The knowledge resource should advance the state-of-the-art in how knowledge representation and reasoning is leveraged in engineered systems.  As there are ongoing initiatives assembling knowledge in data graphs (e.g., Wikidata uses RDF), and ontologies (eg., Stanford's BioPortal uses OWL), we are envisioning that the proposed resource will use a language more expressive than these existing efforts.  Examples of such languages include answer set programming~\cite{lifschitz2019answer} and its query-driven variants such as s(CASP)~\cite{scasp}, Rulelog/Ergo~\cite{ergo+rulelog-prolog50} and other extended well-founded logic programs, theorem proving systems such as Lean~\cite{moura2021lean}, constraint representation~\cite{dechter2003constraint}, methods to incorporate uncertainty~\cite{koller2009probabilistic} and causal information \cite{Pearl-book1,pearl-book2}.

The knowledge resource should foster a distributed community, document use cases, and enable collaboration. Instead of wasting time debating ideal knowledge representations, the community's focus should be on developing shared infrastructure and interoperability, and on maintaining consistency across representations, whenever possible. The resource should enable unanticipated linkages across diverse domains, for example, environmental knowledge could be cross linked to health knowledge; population knowledge could be linked to economic factors; etc.

\subsection{Beneficiaries of a Knowledge Resource}
The knowledge resource must exist in the modern ecosystem of societal needs and challenges. 

The knowledge resource will benefit AI engineers creating  novel AI applications by weaving together different off-the-shelf components. If knowledge modules are made available in a way that engineers can use them as easily as installing a new Python library, a range of AI applications will benefit including those incorporating agentic workflows and generative AI models. 

The knowledge resource will also be an important new form of data for training machine learning systems. Most existing machine learning data sets are limited to facts.

The knowledge resource will benefit applications that require world models. Examples of such applications include those from the robotic planning community which  must bridge between sensor-level data and higher-level cognitive world models, and the fact-checking systems that must detect factually false information.

 With the growth in biomedical literature, it is not humanly possible for any one person to absorb and synthesize everything that is being published. A knowledge resource will enable superhuman knowledge aggregation and consistency maximization over complex domains (such as molecular biology, medicine or social policy)~\cite{witbrock2015cyc}. 

A knowledge resource will benefit knowledge representation and reasoning researchers by catalyzing research and application in this area. UC Irvine’s library of machine learning tasks and data sets~\cite{uci_machine_learning_repository}, and the Hugging Face repository of transformer models~\cite{huggingface} have catalyzed the research and adoption in machine learning and natural language processing. A similar open-source resource will benefit 
%agc changed next line to read better
%the researchers on knowledge representation and reasoning.
knowledge representation and reasoning researchers.

\section{Formalizing foundational knowledge}
Establishing foundational knowledge involves representing abstract knowledge such as knowledge about time, space, actions, causality, and mid-level knowledge such as the working of physical devices, or tenets of social psychology, qualitative physics, etc.~\cite{davis2015commonsense}. To achieve maximum reusability and applicability, the knowledge resource must leverage foundational knowledge. We organize our discussion along four dimensions: easily available foundational knowledge, methodology for collecting and using foundational knowledge, evaluating its impact on the system behavior, and some long-term challenges.

\subsection{Easily available foundational knowledge}

As there has been much work in representing foundational knowledge, a knowledge resource can easily bootstrap from this prior work. This includes qualitative representations of time and space~\cite{cohn2008qualitative,ASP-Space-Non-Monoto-2015,forbus2019qualitative}, representations of events and actions~\cite{gelfond1998action}, formalizations of psychology~\cite{gordon2017formal}, and representations to capture constraints~\cite{dechter2003constraint} and probablistic knowledge~\cite{Pearl-book1,pearl-book2,koller2009probabilistic,darwiche2009modeling}.  OpenCyc’s ontology with integrated representations for vision, space, and language is also available~\cite{NextKBCite}.
           
\subsection{Methodology for collecting foundational knowledge}

The success of some past foundational knowledge representation efforts can be attributed to the rare contributions of ontological prodigies, for example, Pat Hayes, Jerry Hobbs and Ernest Davis. In contrast, a scalable approach requires community-curated resources that leverage repeatable processes and well-defined engineering best practices for building foundational knowledge infrastructure. To support this, a structured, multi-step approach is desirable.

First, start with a broad theoretical understanding by identifying key concepts in a domain. This process can be aided with corpus analysis, for example, by identifying sample texts in a domain for the occurrence of certain concepts~\cite{chaudhri2014creating}. 

Second, the dual goals of broad coverage and inferential competency in foundational theories can be pursued via the approach of successive formalization~\cite{gordon2017formal}. In this method, first-draft axioms are authored to support inference across the entire set of domain concepts previously identified in the first step, then incrementally formalized into competent logical theories through elaborations and refinements. The process of representation is iterative in that the initial choices for predicates and functions will likely evolve as the work progresses.  It is  crucial to recognize both obvious and subtle issues and to balance theoretical perspectives with practical problem-solving. Through this iterative process, one can develop a more refined understanding.

Finally, large language models can potentially contribute to the creation of foundational knowledge. This is, however, an area of research that is under-explored~\cite{llmskg2024}. An initial discussion on the use of LLMs appears in Section~\ref{llms-for-curation}.

\subsection{Evaluating foundational knowledge}

We can evaluate foundational knowledge using both intrinsic and extrinsic methods. Intrinsic evaluation involves a theoretical analysis based on criteria such as completeness, consistency, elaboration tolerance, redundancy, etc.  Extrinsic evaluation involves showing that the use of foundational knowledge improves performance on a suite of tasks. Extrinsic evaluation is more easily understood by the stakeholders because of its direct connection to specific problems. Foundational knowledge can also be evaluated on the basis of whether it can inform the creation of nuanced and efficient test datasets. For example, instead of creating a broad test data set in which all examples are similar, we can use the foundational knowledge to identify the corner cases, and then design the test data to cover those corner cases, thus yielding a more compact test set. Finally, an indirect measure of the usefulness of foundational knowledge is in its  adoption and reuse beyond its original creators.

\subsection{Long-term challenges in formalizing foundational knowledge}

There is no dearth of problems when it comes to formalizing foundational knowledge, but a few of them stand out. 

First, real-world problems tend to be multi-modal in nature. For example, foundational knowledge will need to handle text combined with images, or mix audio, video and text.  The foundational knowledge must address conceptual, temporal, and spatial aspects taken together.  

Second, there needs to be a better bridge between the foundational knowledge and the abstract and vague aspects of natural language. Different logical forms of verbs, and different senses of words, can be typically understood by humans through surrounding context, but the current foundational theories offer no similar mechanisms. For example, consider the sentences "the bottle contains wine" and "the wine contains alcohol". Foundational theories must provide a way to disambiguate between such different uses of "contains".

Finally, there is still no easy way to translate across different representations of knowledge. For storing pictures, for example, there are multiple formats such as JPEG, PNG, etc. Translation tools exist for going across them, although some of the translations can be lossy. In contrast, knowledge represented in one particular formalism remains locked into that formalism, and it is not straightforward to exploit it in a system different from the one it was originally developed in. Creating a standard for interoperability between formalisms, such as was done for the less expressive language OWL, is a possible approach.

\section{Automated Reasoning}
The word ``reasoning'' has been used to refer to a variety of computational processes~\cite{aaai2025panel}. On one hand, we have deductive or probabilistic reasoning in which typically there is a formal proof that relates a question to its answer. On the other hand, we have inductive reasoning and analogical reasoning in which there may not always exist a formal proof that relates a question to an answer. In contrast to classical forms of deductive and inductive reasoning, a major emphasis in logic and knowledge representation and reasoning (KR) research has also been on abductive reasoning, which is a cornerstone of hypothesis formation and belief revision. Computer scientists strive to associate formal properties and guarantees for all of these reasoning processes. Examples of such formal properties include soundness, completeness and tractability.  

For the purpose of the present paper,  the term reasoning refers to the full spectrum of computational processes that have been assigned this label in the literature.
%e.g., encompassing aspects reasoning about space, motion, actions, events, and change.  -- This does not quite fit here.
Given this broad notion, we will organize our discussion along the following dimensions: discovering axioms automatically, modeling  human reasoning, reasoning at scale, and incorporating context into reasoning.

\subsection{Discovering Axioms}

The problem of automatically discovering axioms has been traditionally studied under the topic of inductive logic programming~\cite{muggleton1991inductive}. Related efforts exist in qualitative reasoning to learn new patterns based on analogical generalization~\cite{mclure2010learning} or as default rules~\cite{foldse}. Modern practice, however, relies on a combination of manual and automatic approaches. For example, while pursuing Bayesian causal reasoning, it is expected that partial knowledge about the world is provided by an external source. Likewise, inductively acquiring knowledge about everyday embodied human interactions (e.g., from multimodal data) requires support for specialized domains such as space, time, and motion~\cite{ILP-learning-space-time}.  To support reliable and predictable decision support, large language models are coupled with an external module that contains human verifiable and explicit knowledge.

\subsection{Open challenges in modeling human reasoning}

Human reasoning is often prone to error. For instance, people frequently favor conclusions that are psychologically appealing over those that are mathematically sound~\cite{tversky1974judgment}. This raises concerns about designing AI systems that directly mimic human reasoning as they risk inheriting these flaws. A more pragmatic objective is to develop AI systems that assist and augment human reasoning, helping individuals arrive at more accurate and reliable conclusions. In the following section, we outline the key challenges in pursuing this goal.

First, it will be helpful to develop a taxonomy of distinct kinds of human reasoning.  Such a taxonomy will enable better communication about which aspect of human reasoning is being modeled in a computational system. A partial  taxonomy is available in the existing literature for reasoning tasks such as query answering, planning, projection, diagnostics, etc.  

Second,  the human understanding of the world has a symbolic structure, and AI programs must exploit this to reason correctly even though they might rely on raw data for some of the processing. For example, when humans engage in diagrammatic reasoning, they are able to supplement their purely logical reasoning with diagrams and sketches by exploiting the symbolic structure of the world.

Third, just as humans can resolve ambiguities and conflicts during a conversation, reasoning tools must be able to do the same. Humans often make judgment calls because of these ambiguities and conflicts. Reasoning tools should be able to represent such judgment calls.

Fourth, the reasoning systems must be good at ignoring irrelevant details.  For example, in traditional procedural programming a local variable has effect only in a certain scope. Similarly, these reasoning processes should be able to keep only a subset of facts in scope that are relevant for the current problem. 

Fifth, the reasoning tools should be such that complex reasoning mechanisms that humans use, such as deduction, abduction, induction, counterfactual reasoning, etc., can be elegantly captured.

Finally, graphical models, be them deterministic or probabilistic enable causal and counterfactual reasoning that are central to human reasoning. A primary challenge is to acquire the causal model, even partially (namely the causal graph). Once a partial causal model is available, computing causal effect and counterfactual reasoning will be facilitated and should be further explored.

\subsection{Reasoning at scale}
With the growing knowledge in science, it is humanly impossible for any one person to effectively reason with it all at once. Reasoning at the scale of all science is, therefore, a practically useful challenge for AI systems. Reasoning can be particularly effective in processing what is already known about science, and pinpointing gaps for further research.  For example, structural causal models and constraint reasoning models can be especially effective to support drug discovery and protein design. 

\subsection{Incorporating context into reasoning}

Incorporating real-world constraints into reasoning is necessary for it to work correctly. Cyc's knowledge base achieved this goal by organizing its knowledge into microtheories  into a hierarchical structure~\cite{lenat1995cyc}. The same goal can be achieved through other methods. 

We must not assume our knowledge to be a single monolithic structure.  Different knowledge modules that apply to different contexts should be able to interoperate with each other depending on the problem at hand. 

Real-world reasoning scenarios themselves present constraints that the reasoning process should be able to pick up. For example, designing an artifact requires understanding its operating conditions to ascertain what materials are appropriate. 
% As another example, a video generation or interpretation application should understand that that people do not randomly disappear from a scene, or that occluded objects do not cease to exist~\cite{out-of-sight-ijcai2019}. 
% Above example commented out base don this comment: (Certainly important, but is this a matter of context? Seems more like background knowledge. I guess all these kinds of additional knowledge can be counted as context. But in most discussions of `context' that I have been involved in, we were mostly thinking in terms of how the same vocabulary and/or propositions might have different interpretations (and hence obey different axioms) in different circumstances. Of course, modular organisation of knowledge into different domains or types of scenario can also account for shifts of meaning of terms, since they can be localised to a particular module (eg formulated as a micro-theory)) 
Many of these issues can be addressed by checking inconsistencies with the real-world constraints, and through the development of specialized but domain-independent solvers integrating aspects such space, motion, actions, events, and dynamics~\cite{ASP-Space-Non-Monoto-2015,SEMANTIC-Q-A-CLP-2016}.

As science is a social process, reasoning methods must gracefully integrate with existing workflows and take into account the assumptions that the scientists are making. For example, a recent experiment combining qualitative process theory with a language model led to substantial improvements in performance~\cite{barresMcFateFromBuilding}.

Finally, as much of reasoning in science adopts a certain point of view, choosing the correct viewpoint is crucial. As an example, the shape of the Earth is viewed differently in topography versus astronomy.

\section{Knowledge Curation using manual methods, machine learning and Large Language Models }

Knowledge curation is the process of gathering, extending, and maintaining knowledge.   Knowledge curation encompasses the full life cycle of specifying the requirements, schema design, data cleaning and loading, debugging and troubleshooting, and revising the knowledge.

Current practice on knowledge curation relies on teams of knowledge engineers and  domain experts.  We believe that this needs to change. Next we address the role of human oversight and automation in knowledge curation, and popularizing knowledge curation among scientific communities.

\subsection{Role of human oversight in knowledge curation}

Human oversight is indispensable in any knowledge curation effort. We will illustrate this using three use cases: knowledge curation to support web search, compliance to emission standards, and development of machine learning solutions.

Accurate results for web search require durable semantics. For example, a movie can go through several stages — initial publication of the story as a novel, availability of screenplay, filming and production, box office release, Netflix release, release in other languages, etc. As this process unfolds over a number of years, the search engine must correctly correlate different versions of the movie. At present, such correlation requires human oversight through a careful schema design.

To test the compliance of emission standards of automobiles, much sensor data is available, but most of it is not relevant. Human oversight is needed to identify the relevant aspects of sensor information that should be used for checking compliance to standards.
    
Most machine learning approaches require data that must be annotated by humans. Once the model is trained, \textit{reinforcement learning with human feedback}  is a vital part of model fine-tuning.  Once a machine learning model is deployed, human oversight is necessary to ensure that the model performance does not drift as the input data evolves.

In summary, human experts can provide initial domain knowledge artifacts and define their intent to guide automated knowledge curation systems. This human-guided initialization helps systems understand specific domain contexts and requirements before any automation begins. Even after the initial automated generation of logical structures, human reviewers must validate and refine generated representations, ensuring accuracy and alignment \cite{Akinfaderin_Diallo_2025}. The quality of automatically curated knowledge ultimately depends on human expertise to verify that automated outputs correctly represent the intended domain knowledge.

\subsection{Role of Large Language Models in knowledge curation}
\label{llms-for-curation}
We assume that any step of the knowledge curation process that could be automated should be automated. The question we address is whether previously human labor-intensive tasks can now be automated with the advent of LLMs?  LLMs can enable knowledge curation in at least three ways: becoming a source of knowledge, aiding in knowledge elicitation, and serving as knowledge curators.
	
LLMs capture an immense amount of knowledge implicitly. We can interrogate them to explicitly emit their knowledge on topics of interest for a given application.  Such explicit knowledge, either in a natural or a formal language, can be used in multiple ways. It can be directly built into the application under human oversight and used by a reasoning process. It can also be used as a way to gain insights into the domain of interest which can speed up the downstream design work of curators.
	
LLMs have enabled the construction of powerful chat bots. This capability could be leveraged towards creating a systematic methodology to facilitate interdisciplinary knowledge acquisition. LLM-based natural language dialogs would need to be designed that support a domain expert in articulating knowledge which can be further curated by knowledge engineers or by the LLM itself.
	
LLMs might be used  as components in knowledge curation  interfaces between humans and large complex knowledge resources. As knowledge resources get big, they become difficult to understand by casual users.  LLMs could provide an ability for a broader class of users to add and contribute their knowledge to a knowledge resource. In this scenario, an AI system using an LLM is a useful mediator between a human and a complex knowledge base.

In practice, frontier knowledge curation systems powered by LLMs have shown promise in analyzing documents, identifying key concepts, translating natural language into formal representations, and combining them into comprehensive knowledge models \cite{Akinfaderin_Diallo_2025}. This automation significantly reduces the manual effort traditionally required for knowledge formalization. These systems can auto-formalize in order to validate claims, applying automated reasoning to detect factual inaccuracies with minimal human intervention and with explainability built-in. When validation fails, advanced systems can generate suggestions showing alternative representations that would resolve inconsistencies, effectively automating parts of the knowledge refinement process that previously required extensive human expertise.

We posit that an optimal approach to knowledge curation combines human expertise with automated systems in a continuous feedback loop. In this paradigm, humans provide initial knowledge artifacts and domain expertise, while automation handles complex transformations into formal structures that support verification \cite{Akinfaderin_Diallo_2025}. Human-in-the-loop testing and validation exemplify this partnership: humans review and pose test scenarios and evaluate presented outcomes, while automated systems apply rigorous validation against established, formalized knowledge bases. When inconsistencies are detected, these systems provide grounded explanations and suggestions, which humans can then use to refine knowledge representations or improve system responses. This creates a continuous improvement cycle where human oversight guides automation, and automation enhances human capabilities.

\subsection{Popularizing knowledge curation among scientific communities}

The knowledge curation enterprise must be popularized among the scientific computing communities. Database curation efforts already exist across multiple sciences, by one count, there are 27 databases in materials science and chemistry alone~\cite{blaiszik_awesome_matchem_datasets}. But, few scientific communities are currently creating knowledge bases using expressive knowledge representation languages. Though some work leverages knowledge graphs~\cite{seglerModellingChemicalReasoning2017, mrdjenovichPropnetKnowledgeGraph2020}, current approaches have many limitations. For example, the representation of relationships between equations, variables, and broader theories that we see in Wikidata is limited. For instance, in Wikidata, ``Stoke's Theorem" is a ``generalization of" ``Green's Theorem'' -- this captures some connections, but ''generalization of'' carries no deeper meaning about the nature of this generalization. Languages such as Lean provide greater expressiveness for formalizing scientific knowledge~\cite{bobbinFormalizingChemicalPhysics2024, tooby-smithHepLeanDigitalisingHigh2025}, but its complexity makes it challenging to use by users outside formal logic disciplines. 

Nonetheless, projects like PhysLean aim ``to create a library of digitalized physics results in the theorem prover Lean 4, in a way which is useful to the broad physics community''~\cite{physlean}. We imagine formalized versions of scientific texts like Feynman’s Lectures~\cite{feynmanlectures}, where the scientific content and derivations are structured, functional, and executable, with concepts interlinked across the text. Achieving this requires popularizing knowledge curation among scientific computing communities so that more of these specialists come forward to contribute. 

\section{Modern Education on Knowledge Representation}

As highlighted in a recent report of a Dagstuhl seminar~\cite{delgrande2023current}, there has been a consistent decline in the open academic positions in knowledge representation and reasoning, as well as in the number of students
and researchers attracted to this field. There is a concern that after the current faculty members teaching knowledge representation retire, there is no plan to replace them. In this section, we consider in more detail the current practice for teaching knowledge representation, identify what is missing, and outline potential steps for the future. 

\subsection{Current practice for teaching Knowledge Representation}

At most universities, especially in the United States, knowledge representation is taught as part of either an AI course or as a module on logic embedded in a course on discrete mathematics. Some universities provide knowledge representation and reasoning courses as advanced electives, and there are several textbooks to support such courses~\cite{brachman2004knowledge,reiter2001knowledge,hendler2020semantic}.  Currently used standard AI textbooks~\cite{russell2016artificial,poole2010artificial} have several chapters on knowledge representation.

	Several modern textbooks are available on logic programming~\cite{genesereth2022introduction,lifschitz2019answer,gelfond2014answer,gebseretal2012asppractice,darwiche2009modeling}. Three of these textbooks focus exclusively on answer set programming~\cite{lifschitz2019answer,gelfond2014answer,gebseretal2012asppractice}. They are used in courses taught by faculty members associated within that field. 
    Two of these books come with online repositories containing slides and other teaching materials~\cite{onlineasp,potsdamclass}.
    There is also a textbook that  focused on teaching an audience without technical background how to think using computational ideas~\cite{kowalski2011computational} which has found a home in some philosophy courses. In addition, Computer Science departments at many universities offer courses in computational logic that cover knowledge representation and reasoning.
    
	Wright State University has created an educational hub to teach people about knowledge graphs~\cite{kastle_lab_education_gateway}. Their curriculum is tailored to different audiences ranging from students to senior executives. They are also developing an industry certification for knowledge graph professionals in cooperation with the Knowledge Graphs Conference.

    Northwestern University has a Knowledge Representation and Reasoning course that exposes students to logic, Semantic Web technologies, and Cyc-style knowledge bases.  Students get hands-on experience with industrial-scale knowledge bases with a heavy project-based component.
    
	Within the industry, especially for the Cyc KB, teaching materials tailored to the technology have been developed. This teaching material typically assumes familiarity with logic. and focuses on teaching practical skills for expressing a given piece of knowledge formally.

\subsection{Deficiencies in the current education practice}

As recent AI research has been dominated by approaches based on machine learning, the coverage of knowledge representation in standard textbooks~\cite{russell2016artificial,poole2010artificial} is out of date. There is also a tendency to portray the topic in less than positive terms.

The current teaching practice on knowledge representation does not always succeed in conveying its importance for building reliable computing systems.  Most courses, with a few exceptions,  are limited to using small theoretical examples, without making adequate connections to the real-world problems and actual impact. There is inadequate integration of logic into other computer science courses. For example,  many students do not realize that they could use propositional logic to debug their if-then-else statements.

Another major missing piece in the current teaching practice on knowledge representation is in cultivating an ability to identify implicit and explicit knowledge and rules of thumb that capture how the world works. In other words, the current teaching fails to cultivate skills to conceptually model a task domain  and answering questions about what that task/domain is. For example, given a short story such as: “For sale: baby shoes, never worn”, one should be able to infer the implicit possibility of the death of the child and the tragedy of having to sell the shoes. Such skills are cultivated in courses on philosophy and literature, but similar skills are needed for computer science students to be effective at knowledge engineering.

There is also a philosophical deficiency in the framing of computer science curricula as they primarily focus on “how to build” vs “how to understand”.  This encourages students to rush towards coding solutions instead of developing clear specifications.  Clear specifications require them to think formally about the requirements and clarify any implicit information before engineering a solution.  

\subsection{Steps to improve the teaching of Knowledge Representation}
\label{improve-kr-education}
Knowledge Representation and Reasoning community should undertake an initiative 
directed to make the teaching materials easily available. Much can be learned from a similar effort undertaken at the University of California at Berkeley for a course on introduction to AI~\cite{berkAI}. 

% There is a plethora of knowledge representation languages, approaches, domains of interest, and reasoning tools.
%
To make the teaching materials easy to use, the community should develop teaching modules that can be easily picked up and plugged into a variety of  computing courses. The teaching modules should touch on various aspects of knowledge representation, and be accompanied by slides, worked out examples, sample exercises, sample projects, and exam questions. 

Consistent with these goals, the Prolog Education Group (PEG) was founded in 2022, on the occasion of the 50th anniversary of the programming language Prolog, to ``teach logic, programming, sound reasoning, and AI'' to people of all ages and to develop and provide relevant educational resources~\cite{peg}. The efforts of the PEG group need to be expanded beyond teaching programming to include topics of knowledge representation.

Modern platforms 
for instructors such as Gradescope~\cite{gradescope} or GitHub Classroom~\cite{githubclass}
support means to implement automatic grading utilities for programming exercises and projects. Providing such utilities is indispensable for easy adaptation of teaching materials.

Similarly, providing user-friendly sandbox environments online for various modules will lower the barrier for integrating material at different academic levels. For example, while undergraduate students in a computer science program might be asked to install a specific reasoner or solver to execute sample code, an online sandbox environment supported by that reasoner could allow a high school student to practice the same task. 

    New materials need to be developed to address “why care” questions that illustrate the practical application of knowledge representation in multiple fields.  A repository of real-world examples should be developed.

    We should integrate teaching of logic programming into other computing courses. For example, within a course on databases, logic programming should be taught as the foundation of modern database management systems. Similarly, a course on programming languages can incorporate material depicting answer set programming and the algorithmic aspects of systems that support this knowledge representation paradigm. Some universities, for example, University of Nebraska, Omaha, and University of Texas at Dallas, already integrate
    logic programming into other courses as suggested here.

    The community should develop a {\em knowledge representation body of knowledge} similar to {\em software engineering body of knowledge} \cite{swebok2014}, against which certifications could be granted. Teaching materials should also underscore the importance of interoperability, demonstrating how different knowledge representations and reasoning systems can be integrated. A key component of the teaching materials should be clear instructions and examples of how to integrate the knowledge resources into different applications.

    Much of R\&D in knowledge representation over the last 2--3 decades has focused on extending the expressiveness of declarative logic programs to go beyond that of knowledge graphs and relational databases. AI education should include coverage of those key expressive features, including answer set semantics versus well-founded semantics, higher-order syntax, constraints, etc.

    % at least: default negation; answer set / stable semantics versus well-founded semantics; defeasibility, strong negation, prioritization, and argumentation; higher-order syntax (HiLog), reification, rule identifiers, and annotations; restraint (a form of semantic bounded rationality); probabilistic uncertainty (both Bayesian and non-Bayesian); constraints; linear programming formulations of logic programming; and hybrid learning-and-reasoning approaches.  
    
    AI literacy courses are being developed across many institutions to teach people about new developments. The community should engage with such initiatives to ensure that the role and value of knowledge representation is adequately covered. It is especially important to include an adequate history of different developments to avoid reinvention of already-known concepts.
     
    It would be valuable to introduce set theory and logic in high schools~\cite{scientific_american_logic}.  These students should be taught different forms of reasoning including deduction, abduction and induction.  The importance of logic should be highlighted at an early age through games, puzzles and debates. This will not only make students critical thinkers, but also alleviate any fear of logic courses in college. The Prolog Education Group has initiated efforts in this direction~\cite{logicalthinking}.
	
    Most courses in computer science, except perhaps software engineering, can be re-framed as knowledge representation courses by focusing on objects, their relationships, and the types of questions or tasks associated with them. Even in software engineering, precise reasoning is critical for developing requirement specifications. The key distinction between an introductory programming course and a knowledge representation course lies in the type of knowledge being represented, its intended audience, and its impact. By embedding knowledge representation throughout the curriculum, students could be trained as both computer scientists and implicit knowledge representation experts—without explicitly realizing it. This approach reframes computer science not just as an engineering discipline but also as a natural science.
	
    More broadly, universities can be encouraged to develop an explicit categorization of courses relevant to knowledge representation (perhaps in computer science, philosophy, library science, etc.). Universities could be encouraged to offer a minor concentration in knowledge representation.  A consolidated resource that gathers different courses and their relevance to knowledge representation would make it easier for universities to offer this minor.

\section{Evaluating a Knowledge Resource}

We can think of evaluating a knowledge resource in at least three different ways. First, we can evaluate individual modules, for example, evaluating foundational knowledge, or evaluating how effectively existing knowledge sets are hosted and disseminated. Second, we can evaluate it in terms of how effectively it fosters a community that helps create it. Finally, we can evaluate a knowledge resource as an enabler of AI and its role in the current frontier of AI developments.

This section situates the knowledge resource evaluation in the context of current developments of AI. From that perspective, we will first consider the limitations of the current practice of evaluations in AI, and then consider a few alternative ways to perform better evaluations through expert interviews, virtual environments, and examination of the working of the system.

\subsection{Limitations of the current evaluation practice}

Current evaluation practice in AI suffers from proxy failure—when a measure becomes a target it ceases to be a good measure~\cite{john2024dead}.  With human testing there is an imperfect correlation between a benchmark and the underlying capability that the benchmark is assumed to measure; with AI evaluation this gap is much wider. For example, a program doing well on a multi-state bar exam is hardly suitable to practice law. Furthermore, when benchmarks are the basis for evaluating and comparing AI systems, those benchmarks are subject to corruption pressures, such as training to the test. 

 The Winograd schema challenge~\cite{kocijan2023defeat} was designed to evaluate the ability to use commonsense knowledge to disambiguate pronoun references. The research community has developed programs that do well on the test without incorporating the explicit common sense that the test was originally meant to test.

We next discuss different approaches to address proxy failure.

\subsection{Evaluation through expert interviews}

Evaluation of an AI system by an expert user interacting with it over just a few hours can be much more insightful than the quantitative metrics reported by benchmarks~\cite{cohn2023dialectical}. Unlike the famed Turing test, the evaluator must know that the subject of evaluation is a computer program, and should be given information about its architecture.  If the task requires domain expertise, the evaluation panel can include both an AI expert and a human expert.

\subsection{Evaluation in virtual environments}
Virtual environments, simulated worlds and games can be effective in evaluating targeted aspects of reasoning. For example, the angry birds competition~\cite{renz2015angry} has been effective at exploring reasoning about actions and qualitative reasoning. Game environments and cognitive robotics tasks can also constrain the allowed moves in ways that can explicitly force any successful program to reason.  Recently proposed Gardner test is an example of such a test that situates the rich tradition of General Game Playing competitions in the context of modern generative AI systems \cite{GardnerTest2025}.

\subsection{Evaluation of the functioning of the system}
Benchmarks should test not just the output of the system but the reasoning steps and individual pieces of knowledge that were used in producing the result.  Even though it may be expensive to produce such test sets from scratch, existing resources such as the Cyc knowledge base could be leveraged to generate such tests. Such a resource would significantly enhance the existing collection of problems assembled by the common sense reasoning community

\section{Next steps towards creating a knowledge resource} 

The AAAI workshop \cite{TIKA2025} that is the basis of the present article is part of a larger initiative \cite{NSF:2514820} to formulate and execute a research program to develop a new knowledge resource for AI. The present workshop is the first step towards finding the requirements and building a community. We have already planned three additional workshops that are focused on use cases for the envisioned knowledge resource.  After the AAAI workshop, a subset of the participants formulated a project to develop a proof of concept for the utility of the knowledge resource.  With the wider community feedback, and the results from the proof-of-concept project, we will be in a better place to design the engineering framework and formulate the associated research problems that must be solved.  For the rest of the section, we discuss the recommendations from the workshop participants, plan for the follow on workshops, and the proof of concept feasibility study.

\subsection{Recommendations from the Workshop Participants}

The workshop participants agreed that a meticulously curated knowledge resource along with the necessary tools and methodologies for its effective use is sorely needed.

For creating such a knowledge resource, much can be learned from the Hugging Face repository~\cite{huggingface}. A similar dynamic platform promoting interoperability among various knowledge representations and reasoning systems should be created.  The portal should be organized around specific tasks that can be performed using the knowledge. It should provide a sandbox for trying out different reasoning capabilities without requiring any licenses or difficult installations. The portal should inspire contributions from students taking knowledge representation courses world-wide. The problems addressed by the portal should be of immediate relevance to a cross section of the industry. It should be straightforward to make use of any given package by simply doing "pip install  $\langle$package-name$\rangle$".

While some argued for intrinsic value of a  knowledge resources on its own, standalone knowledge resources already exist, for example, Common Logic Ontology Repository  \cite{gruninger2012modular}, Standard Upper Merged Ontology \cite{niles2001towards} and Basic Formal Ontologies \cite{arp2015building}. These prior existing knowledge resources differ from the Hugging Face model, and the new knowledge resource envisioned here, in that they do not target any specific end-user task. In addition to focusing on specific tasks, the success of Hugging Face can be attributed to providing interoperability between a few prevalent deep learning models, creating a unified API that simplifies training across different model architectures and tasks, and providing a hub for model training and discovery.

The knowledge resource should be positioned as a source of trusted and verifiable knowledge. LLMs should be explored as an initial use case needing a trusted knowledge resource. In addition to LLMs, a few other use cases must be identified. These use cases must span the breadth of knowledge levels/certainty, as well as mission criticality.

To avoid some of the barriers that have prevented Cyc from being widely incorporated in contemporary AI systems across academia and industry, there was an overwhelming consensus that the knowledge resource needs to be open-source and released under a permissive license.

The community also agreed on the need for effective teaching materials in the form of modules that can be easily adapted by others. More details have been discussed in Section~\ref{improve-kr-education}.

Acknowledging the importance of evaluation and validation, our discussion highlighted the need for test sets and benchmarks, potentially developed in collaboration with existing initiatives such as Cyc and MLCommons~\cite{mlcommons}. This would ensure the reliability and accuracy of the knowledge resources, while also mitigating the risk of redundant or inaccurate data proliferation.	 The creation of public test sets, useful for evaluating the performance of different systems, would be a valuable teaching tool.

A strong consensus emerged that the resource should be created and/or managed by a nonprofit foundation, allowing for membership from both academic and commercial entities. This structure would ensure sustainability and broad participation. To kickstart the initiative, a virtual institute was suggested as an initial phase.

\subsection{Follow on Workshops}

We are planning three follow on workshops that are focused on specific use cases that can benefit from a knowledge resource: education, supply chain and computational law. These topics were chosen because there exists prior work to justify further exploration. We briefly describe each of these workshops. 

The workshop on education is aimed at at identifying problems that cannot be solved using LLMs
alone and require knowledge representation to be created. Examples of such problems include skill graphs, precision knowledge tracing, and grounding AI in verifiable knowledge.  The goal of this workshop is to  formulate a large community-driven knowledge graph that would benefit education.

The workshop on supply chain is aimed at addressing critical vulnerabilities in the global supply chain exposed by the pandemic and infrastructure failures.  We will explore several potential supply chain domains (for example, minerals, plastics, water supply, etc.) with an eye towards identifying a domain where the data is easily available, where a global view of the supply chain could be enabled by creating a rich semantic model.

The workshop on computational law is aimed at creating a community of users for a national library of laws that are represented as computer code.  Preliminary work exists in codifying local building codes suggesting that scaling it to a national level will address significant inefficiency in regulatory compliance landscape.

\subsection{Proof of Concept for a Knowledge Resource}

To adapt the Hugging Face model to knowledge modules, the major roadblock is not the design of a portal itself but to identify concrete problems and use cases where the use of a knowledge resource makes a significant difference. Dr Alessandro Oltramari, the president of the Carnegie Bosch Institute, who attended the workshop, came forward with a set of use cases from Bosch that could be used to establish the value of a knowledge resource. Consequently, Dr. Chaudhri and Prof. Shimizu worked with Bosch to define three use cases: inference gaps in LLMs, robotic skill learning and root cause analysis. We briefly explain each of these use cases.

Through a qualitative reasoning benchmark, Room Space 100, it has been shown that the accuracy of GPT-4 declines from 0.55 to 0.15 when the number of objects in a scene increase from 3 to 6 \cite{li2024reframing}.  In contrast a qualitative spatial reasoner for the same task is complete and correct regardless of the number of users. We will evaluate this claim in the context of a robotic planning use case case provided by Bosch. Bosch currently uses LLMs for both high-level and low-level planning \cite{saxena2024grapheqa}. In our experiment, we will replace the LLM used by high-level planner by a symbolic planner that uses a spatial reasoner, and compare their performance.

Robotic skill learning can benefit by leveraging a knowledge resource. Envision a scenario in which a robot learns to {\em Pick} up a hard object, {\em Orient} it for insertion, and then finally performs the {\em Insert}. In the current approach, a robot would learn this process as one unit. By using a knowledge resource, it could map the process to individual steps such as {\em Pick}, {\em Orient}, etc. which are more generalizable. For example, if a robot has learned to pick up a hard object while doing spark plug insertion, it could use the same skill while moving a hard object from one place to another. Starting from the skill library provided by Bosch \cite{saxena2024grapheqa}, we will evaluate how many of these could be mapped to existing resources to enable generalizable robotic skill learning.

Root cause analysis in manufacturing involves identifying the cause for a defect. For example, temperature spike could be caused by machine vibration which could be caused by an alignment error.  In the current approach, the root cause analysis primarily relies on a machine learning algorithm that processes sensor data.  If some of the causes are modeled in structured knowledge \cite{jaimini2023ontology}, root causes analysis need not rely solely only on sensor data.  We will evaluate how the use of a knowledge resource improves the root cause prediction accuracy.

If the results from any of the above evaluations are positive, that will serve as a template to replicate similar applications to other use cases. Such use cases can also be the starting point for the envisioned open engineering framework.

\section{Conclusion}

It is natural to question the value of a curated knowledge resource in the context of modern AI. The world is captivated by ever-larger generative AI models that produce fluent text, lifelike images, and powerful predictions. These systems dominate the headlines, attract billions in investment, and fuel the race for AI supremacy. No matter how impressive, the generative AI systems have dangerous flaws: they reflect the data they were trained on, lack formal guarantees, and will always have inference gaps.  They offer performance without understanding, fluency without truth. And yet, the incentives of today’s AI ecosystem—benchmarks, funding, and hype—reward more scale, not more sense.

That is why the kinds of knowledge resources pioneered in the classical AI must return. Knowledge resources built on logic, rules, and meaning offer exactly what generative AI lacks: transparency, justification, formal guarantees. They let us encode shared human knowledge and values, instead of outsourcing everything to inscrutable models. Alone, symbolic AI once faltered. But in partnership with deep learning, it can give us systems that are both powerful and trustworthy.

The research culture of deep learning has much to teach us. Abundant tools, open-source frameworks and shared data sets that are the staple of deep learning research make it easy for newcomers to experiment. Symbolic AI knowledge resources lack comparable support and require painstaking modeling and domain expertise. This creates barriers to their adoption by AI engineers.

Our hope through the present workshop, and the follow on activities is to re-energize the knowledge representation and reasoning community in creating knowledge resources that parallel the deep learning models by adapting the engineering practices exemplified by Hugging Face. We must, however, start small and demonstrate the value and effectiveness of knowledge resources within the modern context of Generative AI systems. We have embarked on that journey by identifying industry use cases where LLMs have inference gaps, machine learning is done on very specific skills, and causal reasoning is drowning in data. 
We will evaluate the usefulness of existing knowledge resources in these three contexts, and prototype how the knowledge artifacts can 
be disseminated for widespread use.  We hope that this model of working from uses cases to delivering easily reusable engineering products will inspire others in the knowledge representation and reasoning community to undertake similar efforts in their own spheres.

Much work needs to be done to define actionable research
program. Based on the results of the proof of concept study mentioned in the previous section, we hope to define research projects in
all aspects of creating the knowledge resource: methodologies for creating both foundational and domain-specific knowledge, effective reasoning techniques that scale and take context into account, and in translational techniques that can automatically translate a reasoning task framed in one formalism into another formalism. Enabling seamless distributed development,
especially by domain experts is essential for fostering an effective community.  Last but not the least, we must pay 
attention to considering how to involve contributors from diverse geographic, linguistic, cultural, and institutional backgrounds, including underrepresented or low-resource communities.

AI research has always moved in cycles, with certain paradigms rising and falling in prominence. The remarkable progress we now see in deep learning would not have been possible without the persistence of researchers like Geoffrey Hinton, who championed neural networks long before they were in vogue. Just as neural networks have proven powerful for modeling aspects of human perception and cognition, curated knowledge remains essential for capturing structured, declarative understanding — the kind that underpins reasoning, learning, and communication. Although knowledge curation has receded from the forefront of mainstream AI, its value has not diminished. By investing sustained effort into re-integrating curated knowledge into the AI toolkit, we can unlock new forms of robustness, interpretability, and societal impact — and, once again, broaden the horizons of what AI can achieve.

\section{Acknowledgments}
This work has been supported by grant number 2514820 from the United States National Science Foundation. We are grateful to Prof. Ernest Davis for leading the panel discussion on foundational knowledge at the TIKA workshop.
\bibliographystyle{plain}
%%\bibliography{sample-base,refs}

%%\appendix
\end{document}